\pdfoutput=1
\documentclass[10pt,twocolumn,letterpaper]{article}

\usepackage{cvpr}
\usepackage{times}
\usepackage{epsfig}
\usepackage{graphicx}
\usepackage{amsmath}
\usepackage{amssymb}
\usepackage{amsthm}

\usepackage[pagebackref=true,breaklinks=true,letterpaper=true,colorlinks,bookmarks=false]{hyperref}

\cvprfinalcopy 

\def\cvprPaperID{44} 

\ifcvprfinal\fi
\begin{document}

\title{Continual Learning for Anomaly Detection in Surveillance Videos}

\author{Keval Doshi\\
University of South Florida\\
4202 E Fowler Ave, Tampa, FL 33620\\
{\tt\small kevaldoshi@mail.usf.edu}
\and
Yasin Yilmaz\\
University of South Florida\\
4202 E Fowler Ave, Tampa, FL 33620\\
{\tt\small yasiny@usf.edu}
}

\maketitle

\begin{abstract}
Anomaly detection in surveillance videos has been recently gaining attention. A challenging aspect of high-dimensional applications such as video surveillance is continual learning. While current state-of-the-art deep learning approaches perform well on existing public datasets, they fail to work in a continual learning framework due to computational and storage issues. Furthermore, online decision making is an important but mostly neglected factor in this domain. Motivated by these research gaps, we propose an online anomaly detection method for surveillance videos using transfer learning and continual learning, which in turn significantly reduces the training complexity and provides a mechanism for continually learning from recent data without suffering from catastrophic forgetting. Our proposed algorithm leverages the feature extraction power of neural network-based models for transfer learning, and the continual learning capability of statistical detection methods.
\end{abstract}

\section{Introduction}

The number of closed-circuit television (CCTV) surveillance cameras are estimated to go beyond 1 billion globally by the end of 2021 \cite{videosurveillance}. Particularly, video surveillance is an essential tool with applications in law enforcement, transportation, environmental monitoring, etc. For example, it has become an inseparable part of crime deterrence and investigation, traffic violation detection, and traffic management. However, the monitoring ability of surveillance systems has been unable to keep pace due to the massive volume of streaming video data generated in real-time. This has resulted in a glaring deficiency in the adequate utilization of available surveillance infrastructure and hence there is a pressing need for developing intelligent computer vision algorithms for automatic video anomaly detection. 

Video anomaly detection plays an important role in ensuring safety, security and sometimes prevention of potential catastrophes, hence another critical aspect of a video anomaly detection system is the real-time decision making capability. Events such as traffic accidents, robbery, and fire in remote places require immediate counteractions to be taken promptly, which can be facilitated by the real-time detection of anomalous events. However, online and real-time detection methods have only recently gained interest \cite{mao2019delay}. Also, many methods that claim to be online heavily depend on batch processing of long video segments. For example, \cite{liu2018future,ionescu2019object} perform a normalization step which requires the entire video. Regarding the importance of timely detection in video, as \cite{mao2019delay} argues, the methods should also be evaluated in terms of the average detection delay, in addition to the commonly used metrics such as true positive rate, false positive rate, and area-under-the-curve (AUC).

Although deep neural networks provide superior performance on various machine learning and computer vision tasks, such as object detection \cite{dai2016r}, image classification \cite{krizhevsky2012imagenet}, playing games \cite{silver2017mastering}, image synthesis\cite{reed2016generative}, etc., where sufficiently large and inclusive data sets are available to train on, there is also a significant debate on their shortcomings in terms of interpretability, analyzability, and reliability of their decisions \cite{jiang2018trust}. Recently, statistical and nearest neighbor-based methods are gaining popularity due to their appealing characteristics such as being amenable to performance analysis, computational efficiency, and robustness \cite{chen2018explaining,gu2019statistical}.

A key challenge of anomaly detection in videos is that defining notions of normality and abnormality that encompass all possible nominal and anomalous data patterns are nearly impossible. Thus, for a video anomaly detection framework to work in a practical setting, it is extremely crucial that it is capable of learning \emph{continually} from \emph{a small number of new samples} in an online fashion. However, a vast majority of existing video anomaly detection methods are completely dependent on data-hungry deep neural networks \cite{sultani2018real}. It is well known that naive incremental strategies for continual learning in deep/shallow neural networks suffer from catastrophic forgetting \cite{kirkpatrick2017overcoming}. On the other hand, a cumulative approach would require all previous data to be stored and the model to be retrained on the entire data. This approach quickly becomes infeasible due to computational and storage issues. Thus, preserving previously learned knowledge without re-accessing previous data remains particularly challenging \cite{lomonaco2017core50}. Recent advances in transfer learning have shown that using previously learned knowledge on similar tasks can be useful for solving new ones \cite{lomonaco2016comparing}. Hence, we propose a hybrid use of transfer learning via neural networks and statistical k-nearest neighbor (kNN) decision approach for finding video anomalies with limited training in an online fashion. In summary, our contributions in this paper are as follows:
\begin{itemize}
    \item We leverage \emph{transfer learning} to significantly reduce the training complexity while simultaneously outperforming current state-of-the-art algorithms.
    \item We propose a statistical framework for sequential anomaly detection which is capable of \emph{continual} and \emph{few-shot} learning from videos.
    \item We extensively evaluate our proposed framework on publicly available video anomaly detection datasets and also on a real surveillance camera feed.
\end{itemize}

In Section \ref{s:related}, we review the related literature for anomaly detection in surveillance videos. Section \ref{s:proposed} describes the proposed method, a novel hybrid framework based on neural networks and statistical detection. In Section \ref{s:experiments}, the proposed method is compared in detail with the current state-of-the-art algorithms. Finally, in Section \ref{s:conclusion} some conclusions are drawn, and future research directions are discussed.   
\section{Related Works}
\label{s:related}

A commonly adopted learning technique due to the inherent limitations in the availability of annotated and anomalous instances is semi-supervised anomaly detection, which deals with learning a notion of normality from nominal training videos. Any significant deviation from the learned nominal distribution is then classified as anomalous \cite{cheng2015video,ionescu2019object}. On the other hand, supervised detection methods which train on both nominal and anomalous videos have limited application as obtaining the annotations for training is difficult and laborious. To this end, \cite{sultani2018real} proposes using a deep multiple instance learning (MIL) approach to train on video-level annotated videos, in a weakly supervised manner. Even though training on anomalous videos might enhance the detection capability on similar anomalous events, supervised methods would typically suffer in a realistic setup from unknown/novel anomaly types.

A key component of computer vision problems is the extraction of meaningful features. In video surveillance, the extracted features should capture the difference between the nominal and anomalous events within a video. The selection of features significantly impacts the identifiability of types of anomalous events in video sequences. Early techniques primarily focused on trajectory features \cite{anjum2008multifeature}, limiting their applicability to detection of anomalies related to moving objects and trajectory patterns. For example,  \cite{fu2005similarity} studied detection of abnormal vehicle trajectories such as illegal U-turn. \cite{morais2019learning} extracts human skeleton trajectory patterns, and hence is limited to only the detection of abnormalities in human behavior.

Another class of widely used features in this domain are motion and appearance features. Traditional methods extract the motion direction and magnitude to detect spatiotemporal anomalies \cite{saligrama2012video}. Histogram of optical flow \cite{chaudhry2009histograms,colque2016histograms}, and histogram of oriented gradients \cite{dalal2005histograms} are some other commonly used hand-crafted feature extraction techniques frequently used in the literature. 
The recent literature is dominated by the neural network-based methods \cite{hasan2016learning,hinami2017joint,liu2018future,luo2017revisit,ravanbakhsh2019training,sabokrou2018adversarially,xu2015learning} due to their superior performance \cite{xu2015learning}. Contrary to the hand-crafted feature extraction, neural network-based feature extraction methods \cite{xu2015learning} learn the appearance and motion features by deep neural networks. In \cite{luo2017remembering}, the author utilizes a Convolutional Neural Networks (CNN), and Convolutional Long Short Term Memory (CLSTM) to efficiently learn appearance and motion features, respectively. More recently, Generative Adversarial Networks (GAN) have been gaining popularity as they are able to generate internal scene representations based on a given frame and its optical flow. 

However, there has been a significant ongoing debate of the shortcomings of neural network-based methods in terms of interpretability, analyzability, and reliability of their decisions \cite{jiang2018trust}. Furthermore, it is well known that neural networks are notoriously difficult to train on new data or when few samples of a new class are available, i.e., they struggle with continual learning and few-shot learning. Hence, recently  few-shot learning and continual learning have been studied in the computer vision literature \cite{koch2015siamese,sung2018learning,snell2017prototypical,vinyals2016matching,lomonaco2017core50}. However, not a lot of progress has been made yet in the field of continual learning with applications to video surveillance. Hence, in this work, we primarily compare our continual learning performance with the state-of-the-art video anomaly detection algorithms even though they are not tailored for continual learning.

 \section{Proposed Method}
\label{s:proposed}

\begin{figure*}[th]
\centering
\includegraphics[width=1\textwidth]{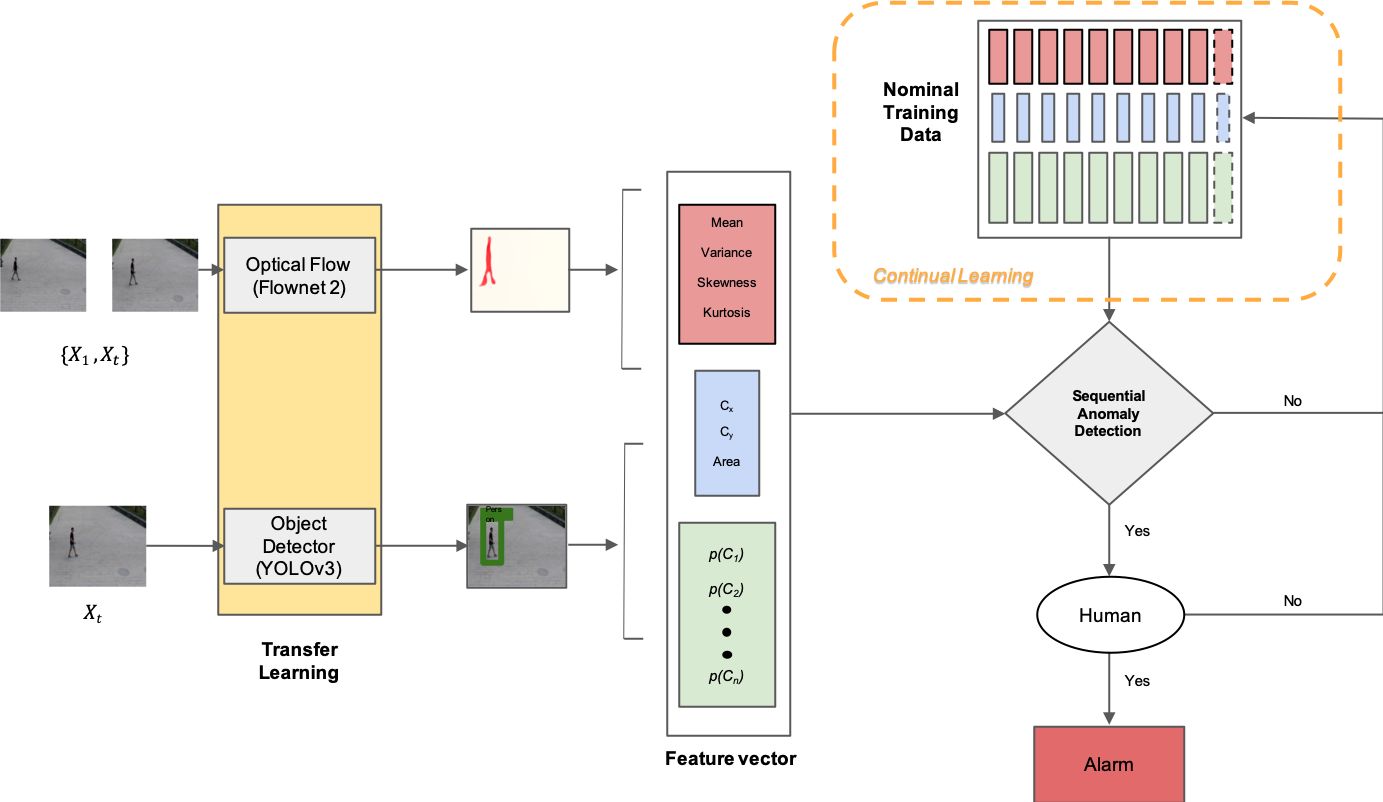}
\vspace{-2mm}
\caption{Proposed continual learning framework. At each time $t$, neural network-based feature extraction module provides motion (optical flow), location (center coordinates and area of bounding box), and appearance (class probabilities) features to the statistical anomaly detection module, which makes online decisions and continual updates to its decision rule.}
\label{f:system}
\vspace{-2mm}
\end{figure*}

\subsection{Motivation}
In existing anomaly detection in surveillance videos literature, an anomaly is construed as an unusual event which does not conform to the learned nominal patterns. However, for practical implementations, it is unrealistic to assume the availability of training data which takes all possible nominal patterns/events into account. Often, anomalous events are circumstantial in nature and it is challenging to distinguish them from nominal events. For example, in many publicly available datasets a previously unseen event such as a person riding a bike is considered as anomalous, yet under different conditions, the same event can be categorized as nominal. Thus, a practical framework should be able to update its definition of nominal events \emph{continually}. This presents a novel challenge to the current approaches mentioned in Section \ref{s:related}, as their decision mechanism is extensively dependent on Deep Neural Networks (DNNs). DNNs typically require the entire training data to be made available prior to the learning task as updating the model on new data necessitates either retraining from scratch , which is computationally expensive, or iteratively with the risk of catastrophic forgetting \cite{kirkpatrick2017overcoming}. Moreover, another motivational fact for us is that the sequential nature of video anomaly detection and the importance of online decision making are not well addressed \cite{mao2019delay}.

\subsection{Feature Selection}

Most existing works focus on a certain aspect of the video such as optical flow, gradient loss or intensity loss. This in turn restrains the existing algorithms to a certain form of anomalous event which is manifested in the considered video aspect. However, in general, the type of anomaly is broad and unknown while training the algorithm. For example, an anomalous event can be justified on the basis of appearance (a person carrying a gun), motion (two people fighting) or location (a person walking on the roadway). To account for all such cases, we create a feature vector $F_t^i$ for each object $i$ in frame $X_t$ at time $t$, where $F_t^i$ is given by $[w_1F_{motion},w_2F_{location},w_3F_{appearance}]$. The weights $w_1, w_2, w_3$ are used to adjust the relative importance of each feature category. 

\subsection{Transfer Learning}
Most existing works propose training specialized data-hungry deep learning models from scratch, however this bounds their applicability to the cases where abundant data is available. Also, the training time required for such models grows exponentially with the size of training data, making them impractical to be deployed in scenarios where the model needs to continually learn. Hence, we propose to leverage transfer learning to extract meaningful features from video. 
\label{s:detect}

\textbf{Object Detection:} 
To obtain location and appearance features, we use a pre-trained object detection system such as You Only Look Once (YOLO) \cite{redmon2016you} to detect objects in video streams in real time. As compared to other state-of-the-art models such as SSD and ResNet, YOLO offers a higher frames-per-second (fps) processing while providing better accuracy. For online anomaly detection, speed is a critical factor, and hence we currently prefer YOLOv3 in our implementations. We get a bounding box (location), along with the class probabilities (appearance) for each object detected in frame $X_t$. Instead of simply using the entire bounding box, we monitor the center of the box and its area to obtain the location features. In a test video, objects diverging from the nominal paths and/or belonging to previously unseen classes will help us detect anomalies, as explained in Section \ref{s:anomaly}.

\textbf{Optical Flow:} Apart from spatial information, temporal information is also a critical aspect of videos. Hence, we propose to monitor the contextual motion of different objects in a frame using a pre-trained optical flow model such as Flownet 2 \cite{ilg2017flownet}. We hypothesize that any kind of motion anomaly would alter the probability distribution of the optical flow for the frame. Hence, we extract the mean, variance, and the higher order statistics skewness and kurtosis, which represent asymmetry and sharpness of the probability distribution, respectively. 

\subsection{Feature Vector}
\label{s:feature}

Combining the motion, location, and appearance features, for each object $i$ detected in frame $X_t$, we construct the feature vector
\begin{equation}
F_t^i =  
\begin{bmatrix}
\vspace{-5mm}
\\ w_1 \text{Mean}
\\ w_1 \text{Variance}
\\ w_1 \text{Skewness}
\\ w_1 \text{Kurtosis}
\\ w_2 \text{C}_{\text{x}}
\\ w_2 \text{C}_{\text{y}}
\\ w_2 \text{Area}
\\ w_3 p(C_1)
\\ w_3 p(C_2)
\\ \vdots
\\ w_3 p(C_n)
\end{bmatrix},
\end{equation}
as shown in Fig. \ref{f:system}, where Mean, Variance, Skewness and Kurtosis are extracted from the optical flow; $\text{C}_{\text{x}}, \text{C}_{\text{y}}, \text{Area}$ denote the coordinates of the center of the bounding box and the area of the bounding box from the object detector; and $p(C_1),\ldots,p(C_n)$ are the class probabilities for the detected object. Hence, at any given time $t$, with $n$ denoting the number of possible classes, the dimensionality of the feature vector is given by $m=n+7$. 


\subsection{Anomaly Detection}
\label{s:anomaly}

We aim to detect anomalies in streaming videos with minimal detection delays while satisfying a desired false alarm rate. Specifically for video surveillance, we can safely hypothesize that any anomalous event would persist for an unknown period of time. This makes the problem suitable for a sequential anomaly detection framework \cite{basseville1993detection}. However, since we have no prior knowledge about the anomalous event that might occur in a video, traditional parametric algorithms which require probabilistic models and data for both nominal and anomalous cases cannot be used directly. Thus, we propose the following nonparametric sequential anomaly detection algorithm. 

\textbf{Training:} Given a set of $N$ training videos ${\mathcal{V} \triangleq \{v_i : i = 1,2,\dots ,N \}}$ consisting of $P$ frames in total, we leverage the deep learning module of our proposed detector to extract $M$ feature vectors $\mathcal{F}^M=\{F^i\}$ for $M$ detected objects in total such that $M \geq P$. We assume that the training data does not include any anomalies. These $M$ vectors correspond to $M$ points in the nominal data space, distributed according to an unknown complex probability distribution. Our goal here is to learn a nonparametric description of the nominal data distribution. We propose to use the Euclidean $k$ nearest neighbor ($k$NN) distance, which captures the local interactions between nominal data points, to figure out a nominal data pattern due to its attractive traits, such as analyzability, interpretability, and computational efficiency \cite{chen2018explaining,gu2019statistical}. We hypothesize that given the informativeness of extracted motion, location, and appearance features, anomalous instances are expected to lie further away from the nominal manifold defined by $\mathcal{F}^M$. That is, the $k$NN distance of anomalous instances with respect to the nominal data points in $\mathcal{F}^M$ will be statistically higher as compared to the $k$NN distances of nominal data points. The training procedure of our detector is given as follows:

\begin{enumerate}
\item Randomly partition the nominal dataset $\mathcal{F}^M$ into two sets $\mathcal{F}^{M_1}$ and $\mathcal{F}^{M_2}$ such that $M = M_1 + M_2$. 
\item Then, for each point $F^i$ in $\mathcal{F}^{M_1}$, we compute the $k$NN distance $d_i$ with respect to the points in set $\mathcal{F}^{M_2}$. 
\item For a significance level $\alpha$, e.g., $0.05$, the $(1-\alpha)$th percentile $d_\alpha$ of $k$NN distances $\{d_1,\ldots,d_{M_1}\}$ is used as a baseline statistic for computing the anomaly evidence of test instances.
\end{enumerate}

\textbf{Testing:} During the testing phase, for each object $i$ detected at time $t$, the deep learning module constructs the feature vector $F_t^i$ and computes the $k$NN distance $d_t^i$ with respect to the training instances in $\mathcal{F}^{M_2}$. The proposed algorithm then computes the instantaneous frame-level anomaly evidence $\delta_t$:
\begin{equation}
\label{eq:evidence}
    \delta_t = (\max_i\{d_t^i\})^m - d_\alpha^m,
\end{equation}
where $m$ is the dimensionality of feature vector $F_t^i$. Finally, following a CUSUM-like procedure \cite{basseville1993detection} we update the running decision statistic $s_t$ as
\begin{equation}
    s_t = \max\{s_{t-1} + \delta_t,0\}, s_0 = 0.
\end{equation}
For nominal data, $\delta_t$ typically gets negative values, hence the decision statistic $s_t$ hovers around zero; whereas for anomalous data $\delta_t$ is expected to take positive values, and successive positive values of $\delta_t$ will make $s_t$ grow. 
We decide that a video frame is anomalous if the decision statistic $s_t$ exceeds the threshold $h$. After $s_t$ exceeds $h$, we perform some fine tuning to better label video frames as nominal or anomalous. Specifically, we find the frame $s_t$ started to grow, i.e., the last time $s_t=0$ before detection, say $\tau_{start}$. Then, we also determine the frame $s_t$ stops increasing and keeps decreasing for $n$, e.g., $5$, consecutive frames, say $\tau_{end}$. Finally, we label the frames between $\tau_{start}$ and $\tau_{end}$ as anomalous, and continue testing for new anomalies with frame $\tau_{end}+1$ by resetting $s_{\tau_{end}}=0$.

\textbf{Continual Learning:} During testing, if the test statistic $s_t$ at time $t$ is zero, i.e, the feature vector $F^i_t$ is considered nominal, then the feature vector is included in the second nominal training set $\mathcal{F}^{M_2}$. When the statistic $s_t$ crosses the threshold $h$, 
an alarm is raised, signaling that the sequence of frames from $\tau_{start}$ to $t$ have never occurred before in the training data. At this point, we propose a human-in-the-loop approach in which a human expert labels false alarms from time to time. If it is labeled as a false alarm, all vectors $\{F^i_{\tau}\}$ between $\tau_{start}$ and $t$ are added to $\mathcal{F}^{M_2}$ so as to prevent similar future false alarms. Thanks to the $k$NN-based decision rule, such a sequential update enables the proposed framework to continually learn on recent data without the need for retraining from scratch, as opposed to the deep neural network-based decision rules. 

\subsection{Computational Complexity}

In this section we analyze the computational complexity of the sequential anomaly detection module, as well as the average running time of the deep learning module.

\textbf{\emph{Sequential Anomaly Detection:}} The training phase of the proposed anomaly detection algorithm requires the computation of $k$NN distance for each point in $\mathcal{F}^{M_1}$ with respect to each point in $\mathcal{F}^{M_2}$. Therefore, the time complexity of training phase is given by $\mathcal{O}(M_1M_2 m)$. The space complexity of the training phase is $\mathcal{O}(M_2 m)$ since $M_2$ data instances need to be saved for the testing phase. In the testing phase, since we compute the $k$NN distances of a single point to all data points in $\mathcal{F}^{M_2}$, the time complexity is $\mathcal{O}(M_2 m)$. On the other hand, deep learning-based methods need to be retrained from scratch to avoid catastrophic forgetting, which would require them to store the old data as well as the new data. The space complexity of the deep learning-based methods would be $\mathcal{O}(abM_2)$ where $a\times b$ is the resolution of the video, which is typically much larger than $m$. Needles to say, the time complexity of retraining a deep learning-based detector is huge. 

\textbf{\emph{Deep Learning Module:}} The YOLO object detector requires about 12 milliseconds to process a single image. This translates to about 83.33 frames per second. Flownet 2 is able to process about 40 frames per second. Accounting for the sequential anomaly detection pipeline, the entire framework would approximately be able to process 32 frames per second. Hence, the proposed framework can process a surveillance video stream in real-time. We also report the running time for other methods such as 11 fps in \cite{ionescu2019object} and 25 fps in \cite{liu2018future}. The running time can be further improved by using a faster object detector such as YOLOv3-Tiny or a better GPU system. All tests are performed on NVIDIA GeForce RTX 2070 with 8 GB RAM and Intel i7-8700k CPU.

\section{Experiments}
\begin{figure*}[th]
\centering
\includegraphics[width=1\textwidth]{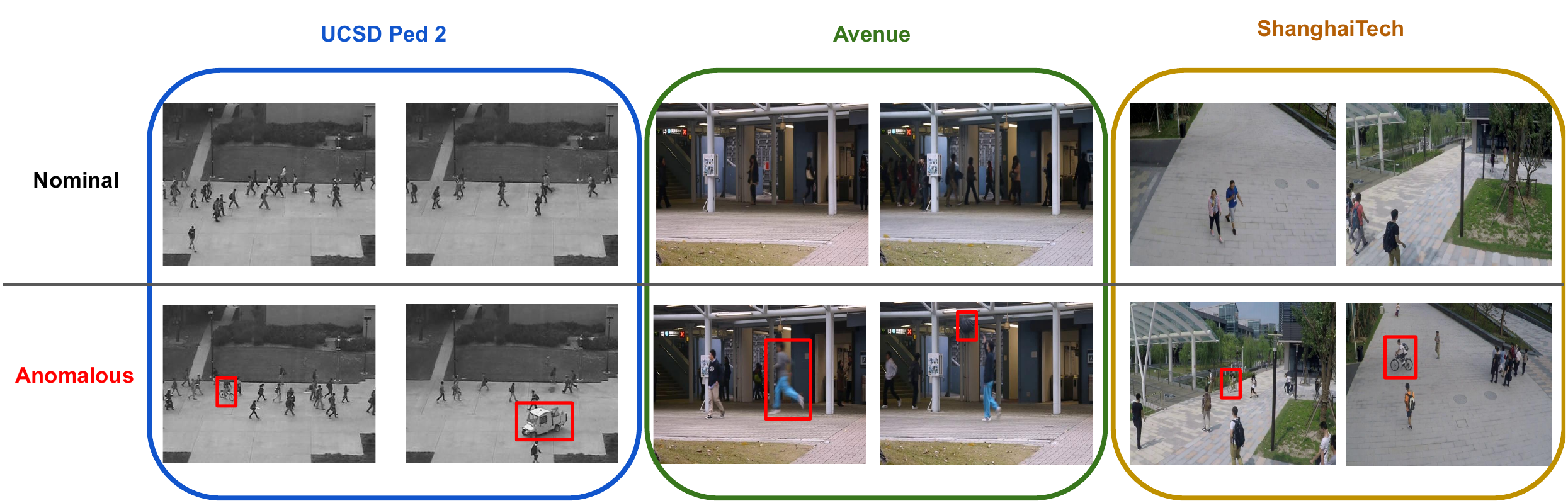}
\vspace{-2mm}
\caption{Examples of nominal and anomalous frames in the UCSD Ped2, CUHK Avenue and ShanghaiTech datasets. Anomalous events are shown with red box.}
\label{f:data}
\vspace{-2mm}
\end{figure*}

\label{s:experiments}

\subsection{Datasets}

We first evaluate our proposed method on three publicly available benchmark video anomaly data sets, namely the CUHK avenue dataset \cite{lu2013abnormal}, the UCSD pedestrian dataset \cite{mahadevan2010anomaly}, and the ShanghaiTech campus dataset \cite{luo2017revisit}. Their training data consists of nominal events only. We present some examples of nominal and anomalous frames in Figure \ref{f:data}. 

\textbf{UCSD Ped2:} The UCSD pedestrian data consists of 16 training and 12 test videos, each with a resolution of 240 x 360. All the anomalous events are caused due to vehicles such as bicycles, skateboarders and wheelchairs crossing pedestrian areas. 

\textbf{Avenue:} The CUHK avenue dataset contains 16 training and 21 test videos with a frame resolution of 360 x 640. The anomalous behaviour is represented by people throwing objects, loitering and running. 

\textbf{ShanghaiTech:} The ShanghaiTech Campus dataset is one of the largest and most challenging datasets available for anomaly detection in videos. It consists of 330 training and 107 test videos from 13 different scenes, which sets it apart from the other available datasets. The resolution for each video frame is 480 x 856. 

\subsection{Benchmark Algorithms}

In the context of video anomaly detection, to the best of our knowledge, there is no benchmark algorithm designed for continual learning. Hence, in Table \ref{tab:my-table}, we compare our proposed algorithm with the state-of-the-art deep learning-based methods, as well as methods based on hand-crafted features: MPPCA \cite{kim2009observe}, MPPC + SFA \cite{mahadevan2010anomaly}, Del et al. \cite{del2016discriminative}, Conv-AE \cite{hasan2016learning}, ConvLSTM-AE \cite{luo2017remembering}, Growing Gas \cite{sun2017online}, Stacked RNN \cite{luo2017revisit}, Deep Generic \cite{hinami2017joint}, GANs \cite{ravanbakhsh2018plug}, Liu et al. \cite{liu2018future}, Sultani et al. \cite{sultani2018real}. A popular metric used for comparison in the anomaly detection literature is the Area under the Curve (AuC) curve. Higher AuC values indicate better performance for an anomaly detection system. Following the existing works \cite{cong2011sparse,ionescu2019object,liu2018future}, we use the commonly used frame-level AuC metric for performance evaluation. 

\subsection{Impact of Sequential Anomaly Detection}

\begin{figure}[th]
\centering
\includegraphics[width=0.5\textwidth]{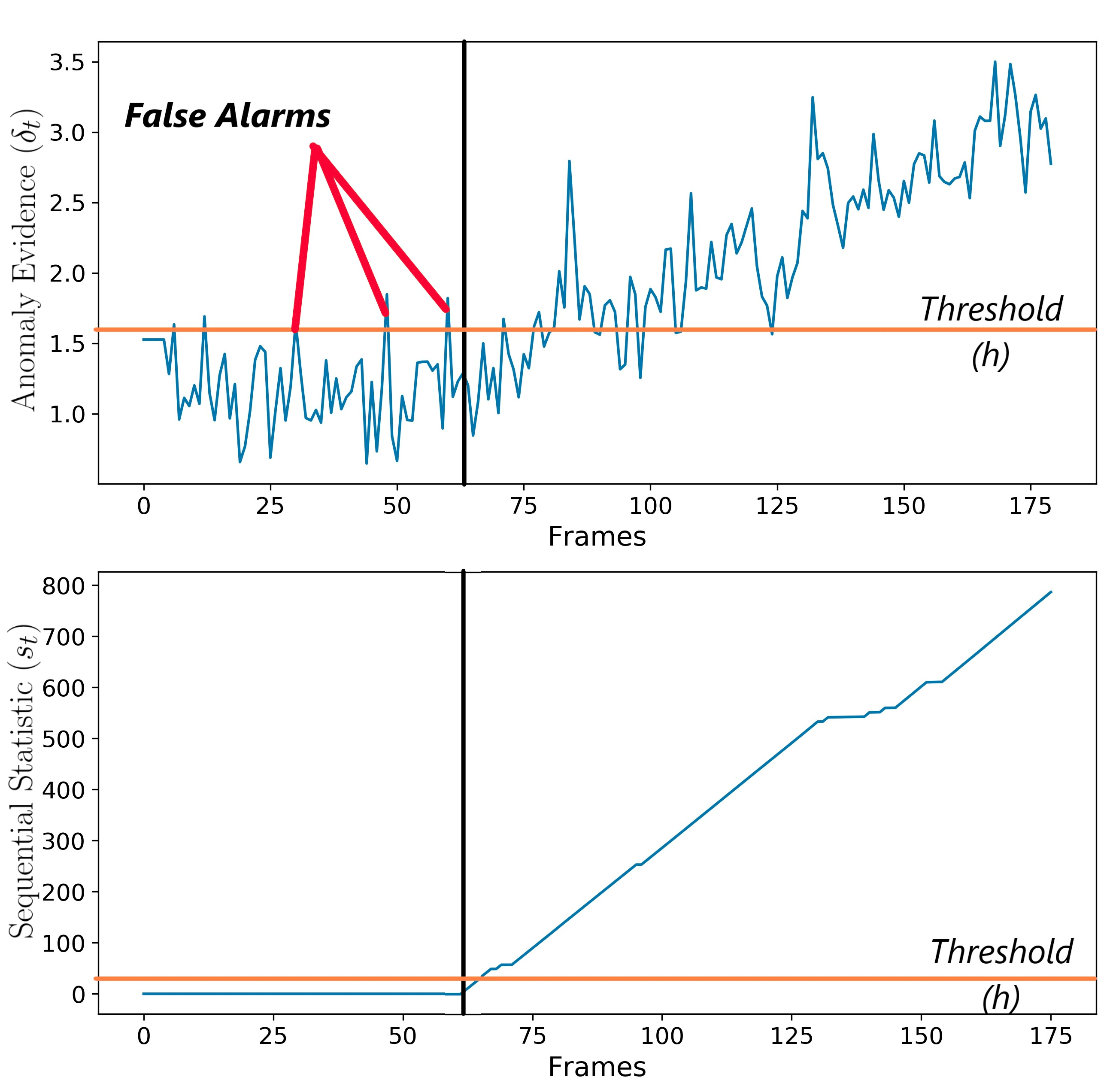}
\vspace{-2mm}
\caption{The advantage of sequential anomaly detection over single-shot detection in terms of controlling false alarms.}
\label{f:sequential}
\vspace{-2mm}
\end{figure}

To demonstrate the importance of sequential anomaly detection in videos, we implement a nonsequential version of our algorithm by applying a threshold to the instantaneous anomaly evidence $\delta_t$, given in \eqref{eq:evidence}, which is similar to the approach employed by many recent works \cite{liu2018future,sultani2018real,ionescu2019object}. As Figure \ref{f:sequential} shows, instantaneous anomaly evidence is more prone to false alarms than the sequential statistic of the proposed framework since it only considers the noisy evidence available at the current time to decide. Whereas, the proposed sequential statistic handles noisy evidence by integrating recent evidence over time. 

\begin{figure}[th]
\centering
\includegraphics[width=0.5\textwidth]{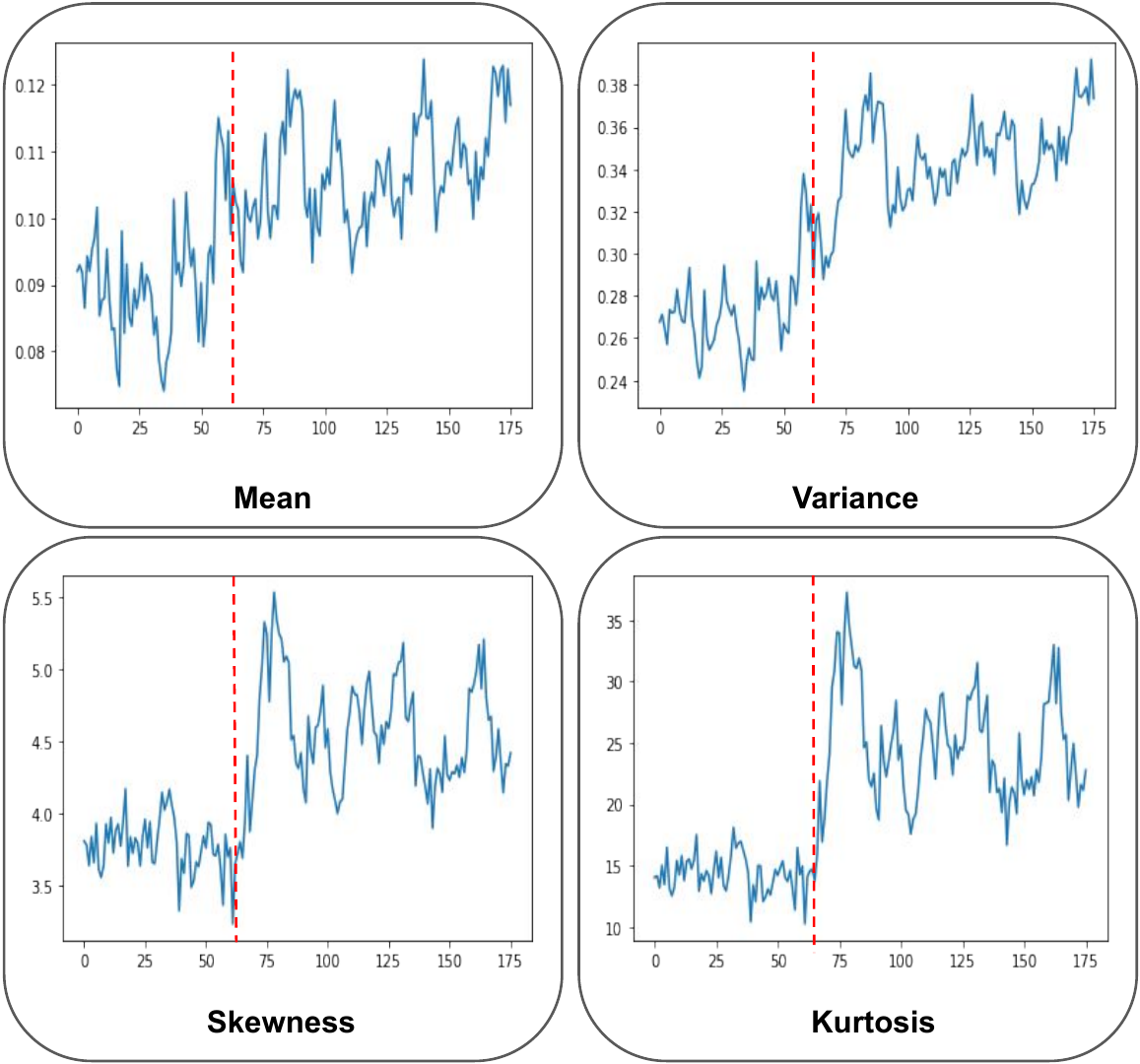}
\vspace{-2mm}
\caption{Optical flow statistics for the motion-based anomaly in the first test video of the UCSD Dataset.}
\label{f:hos}
\vspace{-2mm}
\end{figure}
\subsection{Impact of Optical Flow}
In Figure \ref{f:hos}, we present the optical flow statistics for the first test video of the UCSD dataset. Here, the anomaly pertains to a person using a bike on a pedestrian path, which is previously unseen in the training data. It is clearly visible that there is a significant shift in the optical flow statistics, especially in skewness and kurtosis. This is due to the higher speed of a bike as compared to a person walking. Also, this shows the efficacy of optical flow in detecting motion-based anomalies.     

\subsection{Results}
\textbf{Benchmark Results:}
To show the general performance of the proposed algorithm, not necessarily with continual learning, we compare our results to a wide range of methods in Table \ref{tab:my-table} in terms of the commonly used frame-level AuC metric. Recently, \cite{ionescu2019object} showed significant gains over the rest of the methods. However, their methodology of computing the AuC gives them an unfair advantage as they calculate the AuC for each video in a dataset, and then average them as the AuC of the dataset, as opposed to the other works which concatenate all the videos first and then determine the AuC as the datasets score.

\begin{table}[]
\centering
\resizebox{0.5\textwidth}{!}{%
\begin{tabular}{|c|c|c|c|c|}
\hline
Methodology    & CUHK Avenue & UCSD Ped 2 & ShanghaiTech  \\ \hline
Conv-AE \cite{hasan2016learning}        & 80.0        & 85.0       & 60.9\\ \hline
ConvLSTM-AE\cite{luo2017remembering}    & 77.0        & 88.1       & -   \\ \hline
Stacked RNN\cite{luo2017revisit}    & 81.7        & 92.2       & 68.0\\ \hline
GANs \cite{ravanbakhsh2018plug}           & -           & 88.4       & -   \\ \hline
Liu et al. \cite{liu2018future}     & 85.1        & 95.4       & 72.8\\ \hline
Sultani et al. \cite{sultani2018real} & -           & -          & 71.5\\ \hline
Ours           & 86.4        & 97.8       & 71.62     \\ \hline
\end{tabular}%
}
\vspace{2mm}
\caption{AuC result comparison on three datasets.}
\label{tab:my-table}
\end{table}

As shown in Table \ref{tab:my-table}, we are able to outperform the existing results in the CUHK Avenue and UCSD datasets, and achieve competitive performance in the ShanghaiTech dataset. We should note here that our reported result in the ShanghaiTech dataset is based on online decision making without seeing future video frames.
A common technique used by several recent works such as \cite{liu2018future,ionescu2019object} is to normalize the computed statistic for each test video independently using the future frames. However, this methodology cannot be implemented in an online (real-time) system as it requires prior knowledge about the minimum and maximum values the statistic might take. 

\begin{figure}[th]
\centering
\includegraphics[width=0.5\textwidth]{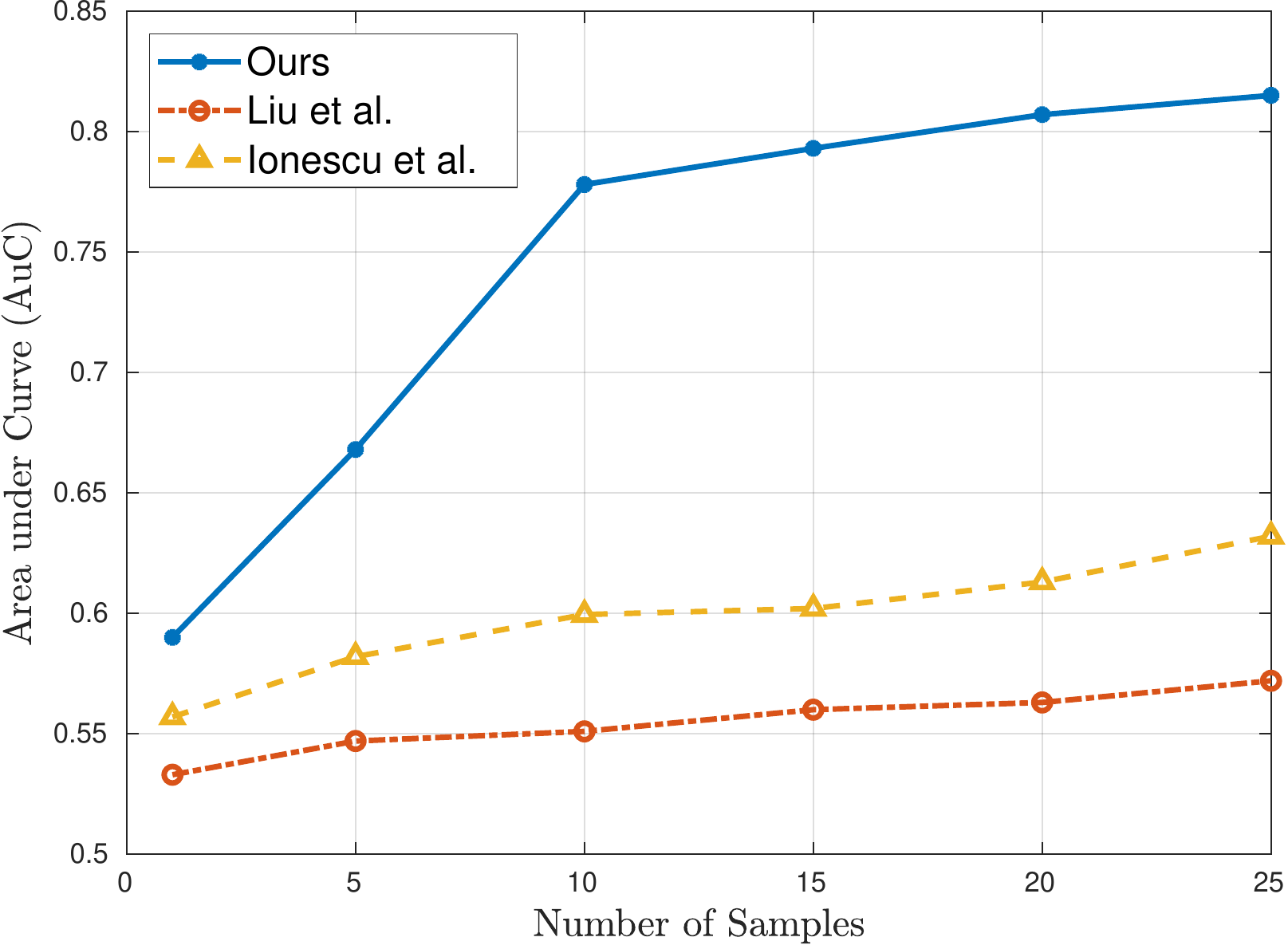}
\vspace{-2mm}
\caption{Comparison of the proposed and the state-of-the-art algorithms Liu et al. \cite{liu2018future} and Ionescu et al. \cite{ionescu2019object} in terms of continual learning capability. The proposed algorithm is able to quickly train with new samples and significantly outperform both of the methods.}
\label{f:continual}
\vspace{-2mm}
\end{figure}

\textbf{Continual Learning Results:}
Due to the lack of existing benchmark datasets for continual learning in surveillance videos, we first slightly modify the original UCSD dataset, where a person riding a bike is considered as anomalous, and assume that it is considered as a nominal behavior. Our goal here is to compare the continual learning capability for video surveillance of the proposed and state-of-the-art algorithms and see how well they adapt to new patterns. Initially, the proposed algorithm raises an alarm when it detects a bike in the testing data. Using the human supervision approach proposed in Section \ref{s:proposed}, the relevant frames are labelled as nominal and added to the training set. In Figure \ref{f:continual}, it is seen that the proposed algorithm clearly outperforms the state-of-the-art algorithms \cite{ionescu2019object,liu2018future} in terms of continual learning performance. More importantly, as shown in Table \ref{tab:time-table}, it achieves this superior performance by quickly updating its training with the new samples in a few seconds while the state-of-the-art algorithms need to retrain on the entire dataset for several hours to prevent catastrophic forgetting. Furthermore, it is important to note that the proposed algorithm is able to achieve a relatively high AuC score using only a few samples, demonstrating its few-shot learning ability.

\begin{table}[h]
\tiny{
\centering
\resizebox{0.5\textwidth}{!}{%
\begin{tabular}{|l|l|l|l|}
\hline

            & Ours                        & Liu et al. \cite{liu2018future}                  & Ionescu et al. \cite{ionescu2019object}              \\ \hline
Update Time & \multicolumn{1}{c|}{10 sec} & \multicolumn{1}{c|}{4.8 hrs} & \multicolumn{1}{c|}{2.5 hrs} \\ \hline
\end{tabular}%
}
\vspace{2mm}
\caption{Time required to update the model for each batch of new samples.}
\label{tab:time-table}
}
\end{table}

\textbf{Real-Time Surveillance Results:}
\begin{figure*}[th]
\centering
\includegraphics[width=1\textwidth]{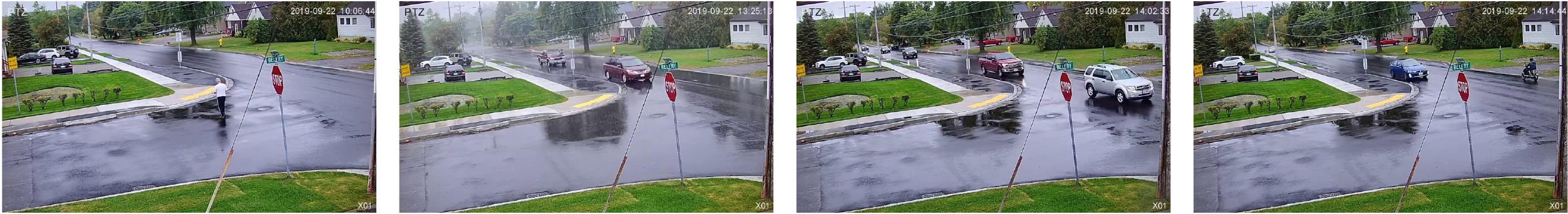}
\vspace{-2mm}
\caption{The visualization of different causes for false alarm in the surveillance feed dataset. In the first case, the person stands in the middle of the street, which causes an alarm as this behavior was previously unseen in the training data. Similarly, in the second case, a change in the weather causes the street sign to move. In the third case, the appearance of multiple cars at the same time causes a shift in the distribution of the optical flow. Finally, in the fourth case a bike is detected, which was not previously seen in the training data.}
\vspace{-2mm}
\end{figure*}
Even though existing datasets such as ShanghaiTech, CUHK Avenue, and UCSD provide a good baseline for comparing video surveillance frameworks, they lack some critical aspects. Firstly, they have an underlying assumption that all nominal events/behaviors are covered by the training data, which might not be the case in a realistic implementation. Secondly, there is an absence of temporal continuity in the test videos, i.e., most videos are only a few minutes long and there is no specific temporal relation between different test videos. Moreover, external factors such as brightness and weather conditions that affect the quality of the images are also absent in the available datasets. Hence, we also evaluate our proposed algorithm on a publicly available CCTV surveillance feed\footnote{The entire surveillance feed is available here: https://www.youtube.com/watch?v=Xyj-7WrEhQw\&t=3460s}. The entire feed is of 8 hours and 23 minutes and continuously monitors a street. To make the problem more challenging we initially train only on 10 minutes of data and then continually update our model as more instances become available. 

\begin{figure}[th]
\centering
\includegraphics[width=0.5\textwidth]{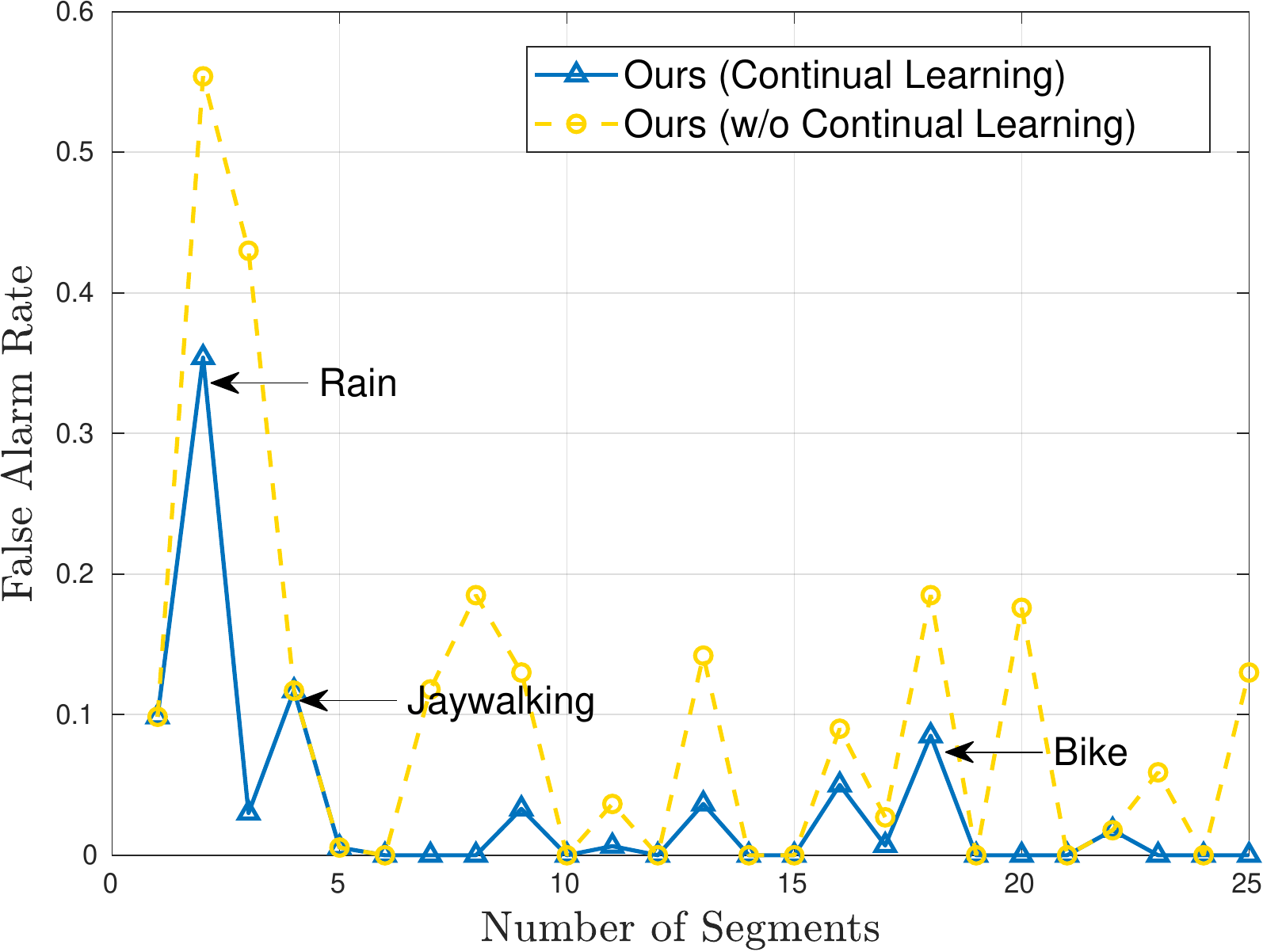}
\vspace{-2mm}
\caption{Continual learning ability of the proposed algorithm. Assuming an arbitrary constant threshold, we observe that the algorithm is able to quickly learn new nominal behaviors, and thus reduce the false alarm rate as compared to the same algorithm which does not continually update the learned model.}
\label{f:cctv}
\vspace{-2mm}
\end{figure}

In Figure \ref{f:cctv}, we demonstrate the continual learning performance of the proposed algorithm through reduced number of false alarms after receiving some new nominal labels.
It should be noted that our goal here is to emphasize the continual learning ability of our algorithm, rather than showing the general detection performance. Each
segment here corresponds to 20,000 frames. After each segment, a human roughly labels the false positive events. In this case, to reduce the computational complexity and examine the few-shot learning ability of the proposed algorithm, we only consider 20\% of all false positive events for updating the model. Although the number of frames might seem a lot, it roughly translates to 10 seconds of streaming video data, so it can still be considered as few-shot learning in video analysis. We observe that even with relatively small updates, the false alarm rate is significantly lower as compared to the same algorithm where we do not update the model continuously. This proves that the proposed algorithm is able to learn meaningful information from recent data using only few samples, and is able to incrementally update the model without accessing the previous training data.   

\section{Conclusion and Future Work}
\label{s:conclusion}

For video anomaly detection, we presented an continual learning algorithm which consists of a transfer learning-based feature extraction module and a statistical decision making module. The first module efficiently minimizes the training complexity and extracts motion, location, and appearance features. The second module is a sequential anomaly detector which is able to incrementally update the learned model within seconds using newly available nominal labels. Through experiments on publicly available data, we showed that the proposed detector significantly outperforms the state-of-the-art algorithms in terms of any-shot learning of new nominal patterns. The continual learning capacity of the proposed algorithm is illustrated on a real-time surveillance stream, as well as a popular benchmark dataset. 

The ability to continually learn and adapt to new scenarios would significantly improve the current video surveillance capabilities. In future, we aim to evolve our framework to work well in more challenging scenarios such as dynamic weather conditions, rotating security cameras and complex temporal relationships. 
Furthermore, we plan to extend the proposed continual learning framework to new anomalous labels, and other video processing tasks such as online object and action recognition.

{\small
\bibliographystyle{ieee_fullname.bst}
\bibliography{Ref.bib}
}
\end{document}